\documentclass[12pt]{article}
\usepackage[utf8]{inputenc}
\usepackage{amsmath}
\usepackage{graphicx,psfrag,epsf}
\usepackage{epsfig}
\usepackage{epstopdf}
\usepackage{enumerate}
\usepackage{natbib}
\usepackage{algpseudocode}
\usepackage{url} 
\usepackage{hyperref, bm}
\usepackage{amsmath,amssymb}
\usepackage{bbm}
\usepackage{xcolor}
\usepackage{caption}
\usepackage{array}
\usepackage{subcaption}
\usepackage{multirow}
\usepackage{amsfonts,amstext,amsmath,amssymb,amsthm}
\usepackage{mathtools}
\usepackage{algorithm}

\newcolumntype{P}[1]{>{\centering\arraybackslash}p{#1}}

\newcommand{\bbR}{\mathbb{R}}
\newcommand{\cW}{\mathcal{W}}
\newcommand{\cV}{\mathcal{V}}

\newcommand{\argmin}{\mathop{\mathrm{argmin}}}

\newcommand{\blind}{0}
\begin{document}

\def\spacingset#1{\renewcommand{\baselinestretch}%
{#1}\small\normalsize} \spacingset{1}


\if0\blind

{
  \title{\bf Variable selection for nonlinear Cox regression model via deep learning}
  \author{Kexuan Li
 \thanks{
    Global Analytics and Data Sciences, Biogen, Cambridge, Massachusetts, USA}}
  \maketitle
} \fi

\if1\blind
{
  \bigskip
  \bigskip
  \bigskip
  \begin{center}
    {\LARGE\bf Title}
\end{center}
  \medskip
} \fi

\bigskip
\begin{abstract}
Variable selection problem for the nonlinear Cox regression model is considered. In survival analysis, one main objective is to identify the covariates that are associated with the risk of experiencing the event of interest. The Cox proportional hazard model is being used extensively in survival analysis in studying the relationship between survival times and covariates, where the model assumes that the covariate has a log-linear effect on the hazard function. However, this linearity assumption may not be satisfied in practice. In order to extract a representative subset of features, various variable selection approaches have been proposed for survival data under the linear Cox model. However, there exists little literature on variable selection for the nonlinear Cox model. To break this gap, we extend the recently developed deep learning-based variable selection model LassoNet to survival data. Simulations are provided to demonstrate the validity and effectiveness of the proposed method. Finally, we apply the proposed methodology to analyze a real data set on diffuse large B-cell lymphoma.
\end{abstract}

\noindent%

\spacingset{1.2}

\newpage
\section{Introduction}
One of the areas of great methodologic advance in clinical trials and medical research has been the ability to handle censored time-to-event data. By censored time-to-event data, we mean that the value of observation is only partly known and the exact time-to-event is not observed. For example, a patient may drop off in the follow-up of the clinical study. For such censored data, we only know that the true time to event is less or greater than, or within the observed time, which refers to left censored, right censored and interval-censored respectively, and thus the classical statistical analysis methods may not be applicable to censored data. Survival analysis is a commonly-used method for the analysis of censored data, where the response is often referred to as a survival time, failure time, or event time. In practice, covariates are collected along with each observation object and the main objective of survival analysis is to demonstrate the dependence of the survival time on these covariates. To achieve this goal, the author in \cite{cox1972regression} introduced a famous semiparametric regression model that calculates the effect of each covariate on the survival time, which is called the Cox proportional hazards model. Instead of modeling the mean function of the survival time, the essential novelty of Cox proportional hazards is to model the hazard function, where the hazard function, as a function of $t$, is defined as instantaneous potential per unit time for the event to occur, given that the individual has survived up to time $t$. To be more specific, the classical Cox proportional hazard model assumes the logarithm of hazard is a linear function of the covariates,  which, however, can be complex and nonlinear in practice. In contrast to the linear Cox proportional hazards model, a considerable interest has also been paid on nonlinear survival models, relaxing the assumption of linearity of log-hazard function (\cite{Fan1997, Random_survival_forests, Deepsurv})

In practice, not all the covariates may contribute to the prediction of the survival outcomes, and researchers are interested in identifying a subset of significant covariates upon which the hazard function depends. For example, in genome-wide association
studies with time-to-event outcomes, people are interested in identifying the diagnostic single nucleotide polymorphisms that are associated with the response. From a statistical point of view, this kind of problem is referred to as variable selection or feature selection. Under the context of classical regression analysis, there are numbers of variable selection techniques, which are roughly categorized into three classes: filter methods, wrapper methods, and embedded methods. In the filter methods, features are filtered based on certain statistical measures for their correlation with the response, such as Pearson's correlation, and fisher score. Wrapper methods evaluate different subsets of covariates on model and the best subset of covariates is selected based on the model performance; some commonly used wrapper methods include forward/backward selection or recursive feature elimination. Finally, in embedded methods, feature selection process is embedded in the learning phase such as the least absolute shrinkage and selection operator (LASSO) (\cite{tibshirani1996regression}), the smoothly clipped absolute deviation (SCAD) (\cite{fan2001variable}), and the adaptive LASSO (ALASSO) (\cite{zou2006adaptive}). The advantage of the embedded method is that the model training and feature selection can be performed simultaneously. See \cite{saeys2007review} for more details on the three methods. Many of these classical variable selection methods have been extended from regression analysis to survival analysis such as best subset variable selection, stepwise deletion, and Bayesian variable selection (\cite{faraggi1998bayesian, ibrahim1999bayesian}). Penalized approach, which optimizes an objective function with a penalty function, has also been extended to right-censored failure time data (\cite{tibshirani1997lasso, fan2002variable, zhang2007adaptive}). More recently, the authors in \cite{fan2010high} extended the iterative sure feature screening procedure to Cox proportional hazards model. The works in \cite{zhao2019simultaneous, du2021unified, yi2020simultaneous} considered the variable selection problem for interval-censored data. However, these methods rely on strong parametric assumptions that are often violated in practice. To overcome this problem, and owing to the advancement of deep learning for handling nonlinearity, we investigate the application of deep neural networks to variable selection for the nonlinear Cox model. To be more specific, we extend LassoNet proposed by \cite{LassoNet} to right-censored survival data.

Besides the above three categories of variable selection methods, in recent years, applying deep learning to variable selection has also made a great breakthrough. To name a few, the work of \cite{liu2017deep, han2018autoencoder} added row-sparse regularization to hidden layers to achieve feature selection; the authors in \cite{abid2019concrete} applied reparametrization trick, or Gumbel-softmax trick to autoencoder and obtain discrete feature selection; in \cite{mirzaei2020deep}, the authors established a teacher-student neural network for feature selection; LassoNet, proposed by \cite{LassoNet}, introduced a novel feature selection framework by penalizing the parameters in the residual layer with the constraint that the norm of the parameters in the first layer is less than the corresponding norm of the parameters in the residual layer; the method introduced in \cite{DeepFS} combined deep neural networks and feature screening under the high-dimensional, low-sample-size setting. Even though these deep learning-based variable selection methods have made great success in regression or classification problems, the extension of them to survival data is not trivial. One of the objectives of this paper is to break this gap.

The rest of the paper is organized as follows. In Section \ref{literiture_review}, we formulate the problem of interest and briefly review the recent applications of deep learning in survival analysis. Section \ref{Method} describes the variable selection procedure in the nonlinear Cox regression model using LassoNet. Section \ref{Simulation} gives several comprehensive simulation studies to evaluate finite sample performances of the proposed methods. Finally, in Section \ref{Real_Data}, we apply our method to diffuse large B-cell lymphoma and give our conclusion in Section \ref{Conclusion}.

\section{Problem Formulation and Related Works} \label{literiture_review}
In this section, we first formulate the problem of interest and then provide a literature review of recent deep learning models applied in survival analysis.

\subsection{Problem Formulation}
Denote $T, C, \bm{X}\in \mathbb{R}^p$ by the survival time, the censoring time, and the corresponding covariates, respectively. For simplicity, we assume that $T$ and $C$ are conditionally independent given $\bm{X}$. In practice, $T$ may not be completely observed, and let $Y=\min\{T, C\}, \delta=\mathbbm{1}(T\leq C)$ be the observed time and the censoring indicator, respectively. Thus, our observed data set is $\big\{(Y_i, \delta_i, \bm{x}_i): Y_i \geq 0, \delta_i \in \{0, 1\}, \bm{x}_i \in \mathbb{R}^p, i=1,\ldots,n \big\}$ which is independently and identically distributed from $(Y, \delta, \bm{X})$. Furthermore, we let $\mathcal{R}(t) = \{i: Y_i\geq t\}$ be the set of all individuals who are available (have survived) at time $t$, which is referred to as the risk set. In survival analysis, with a particular value $\bm{x}$ for the covariate $\bm{X}$, there are two commonly used quantitative measurements to describe the distribution of survival time $T$. The first one is the survival function $S(t|\bm{x})$, which is defined as $S(t|\bm{x}) = \mathbb{P}(T>t|\bm{X}=\bm{x})$, representing the probability of surviving beyond time $t$. Another one is the hazard function $h(t|\bm{x})$, which is defined as:
\[
h(t|\bm{x}) = \lim_{\Delta t \rightarrow0}\frac{\mathbb{P}(t\leq T< T+\Delta t|T\geq t, \bm{X}=\bm{x})}{\Delta t}.
\]
According to the definition of $h(t|\bm{x})$, the hazard function characterizes the
instantaneous rate of failure at time $t$ given that the individual has survived up to time $t$. Given an observed data set $\big\{(Y_i, \delta_i, \bm{x}_i)\big\}$, the corresponding complete likelihood function of the observed data is given by
\begin{equation}\label{complete_likelihood}
L_{\textrm{complete}} = \prod_{i\in\mathcal{U}}f(Y_i|\bm{x}_i)\prod_{i\in \mathcal{ \mathcal{C}}}S(Y_i|\bm{x}_i)=\prod_{i\in\mathcal{U}}h(Y_i|\bm{x}_i)\prod_{i=1}^nS(Y_i|\bm{x}_i),
\end{equation}
where $\mathcal{U}=\{i:\delta_i=1\}$ and $\mathcal{C}=\{i:\delta_i=0\}$ are the index set of uncensored and censored data respectively, and $f(t|\bm{x})$ is the conditional density function. In practice, people are interested in modeling the possible relationship between the survival time $T_i$ and the covariates $\bm{x}_i$. The Cox proportional hazards model in \cite{cox1972regression} is a popular approach to achieve this goal. To be more specific, given features $\bm{x} = (x_1, \ldots, x_p)^\top$, Cox proportional hazards model assumes that the covariate has a log-linear effect on the hazard function, that is,
\begin{equation}\label{CoxPH}
h(t|\bm{x}) = h_0(t)\exp[\bm{x}^\top\bm{\beta}]=h_0(t)\exp[\sum_{j=1}^px_j\beta_j],
\end{equation}
where $h_0(t)$ is a completely unspecified baseline hazard function assuming to be constant across all individuals and $\bm{\beta} = (\beta_1, \ldots, \beta_p)^\top$ is the regression coefficient. The model assumes that the hazard function $h(t|\bm{x})$ is a multiplication of a non-parametric hazard function and a relative risk function, which is given by $\bm{x}^\top\bm{\beta}$. Let $H_0(t) = \int_0^th_0(t)$ be the cumulative baseline hazard function. Under Cox proportional hazards model in (\ref{CoxPH}), using the fact that $S(t|\bm{x}) = \exp[-\int_0^th(t|\bm{x})]$, the complete likelihood function in (\ref{complete_likelihood}) becomes
\begin{equation}\label{complete_likelihood_beta}
L_{\textrm{complete}}(\bm{\beta}) = \prod_{i\in\mathcal{U}}h_0(Y_i)\exp[\bm{x}_i^\top\bm{\beta}]\prod_{i}^n\exp\{-H_0(Y_i)\exp[\bm{x}_i^\top\bm{\beta}]\}.
\end{equation}
Specifically, if $h_0(t)$ (or equivalent $H_0(t)$) is parameterized by $\bm{\theta}$, then both $\bm{\theta}$ and $\bm{\beta}$ can be estimated by maximizing the complete likelihood function in (\ref{complete_likelihood_beta}). However, in practice, the form of $h_0(t)$ is unknown, thus, instead of optimizing $L_{\textrm{complete}}(\bm{\beta})$, $\bm{\beta}$ can be estimated by maximizing the following partial likelihood function, without specification of $h_0(t)$
\begin{equation}\label{CoxPH_loss}
L_{\textrm{partial}}(\bm{\beta}) = \prod_{i=1}^{n}\left(\frac{\exp[\bm{x}_i^\top\bm{\beta}]}{\sum_{j\in \mathcal{R}(Y_i)}\exp[\bm{x}_i^\top\bm{\beta}]}\right)^{\delta_i}.
\end{equation}
From now on, we let $L(\bm{\beta})=L_{\textrm{partial}}(\bm{\beta})$ and $\ell(\bm{\beta}) = \log L(\bm{\beta})$ for simplicity. Notice that the model in (\ref{CoxPH}) assumes the covariate effect on the hazard function is log-linear, however, the linearity assumption is too stringent in real-world applications and thus we should consider more flexible choices for the covariate effect to model the nonlinearity and interaction.

Now, we describe the problem of variable selection in right-censored data. In variable selection, the goal is to identify a subset $\mathcal{S}  \subseteq \{1, 2, \ldots, p\}$ of the most discriminative and informative features with size $|\mathcal{S}| = k \leqslant p$ and also a function $g: [0, \infty)\times\mathbb{R}^k \rightarrow \mathbb{R}$ such that the hazard function $h(t|\bm{x})$ in (\ref{CoxPH}) equals $g(t|\bm{x}^{(\mathcal{S})})$, where $\boldsymbol{x}^{(\mathcal{S})} = (x_{s_1}, \ldots, x_{s_k})^{\top} \in \mathbb{R}^k$ and $s_i \in \mathcal{S}$.

\subsection{Deep Learning for Survival Analysis}
Recently, deep learning has made great breakthroughs in a wide range of applications, such as natural language processing (\cite{Literature_Review_NLP}), computer vision (\cite{Literature_Review_CV}), dynamics system (\cite{ODE}), drug discovery and toxicology (\cite{Literature_Review_drug}). Owing to the superior performance and good theoretical guarantees of deep learning, applying deep learning to survival data has also drawn much attention. In this subsection, we provide a brief review of recent contributions of deep learning in the field of survival analysis.

To the best of our acknowledge, the first attempt to apply deep neural networks in survival data is by \cite{faraggi1995neural}, where the authors extended the linear Cox proportional hazards (\cite{cox1972regression}) by replacing the linear combination of covariates in (\ref{CoxPH}) with a single layer feed-forward neural network. However, they showed that the proposed model fail to outperform the classical linear Cox model. Under a similar framework, instead of using a single-layer neural network, \cite{Deepsurv} explored more complex networks and demonstrated that modern deep learning techniques perform as well as or better than the linear Cox model. In addition, $\ell_1$ and $\ell_2$ regularization terms have been utilized to the loss function to reduce the over-fitting. \cite{kvamme2019time} further alleviated the proportional hazards assumption by approximating the loss function based on case-control sampling. The authors also came up with a heuristic to approximate the gradients, enabling the models to scale to large data sets. \cite{RNN} considered deep recurrent neural network to capture the sequential patterns of the feature over time in survival analysis. \cite{Deephit} introduced a novel approach called DeepHit which is able to directly estimate the joint distribution of events in discrete time. To improve discriminative performance, \cite{Deephit} combined the log-likelihood function with a ranking loss and is applicable for competing risks. \cite{Zhu2017} added convolutional architectures to the loss function in order to capture image-based covariates and applied the method to pathological images of lung cancer. \cite{luck2017deep} combined Cox partial likelihood function with an isotonic regression loss to improve the predictive performance. \cite{manduchi2021deep} studied the problem of clustering survival data by using a deep generative model. \cite{nagpal2021deep} introduced a fully parametric approach, named deep survival machines, to estimate relative risks in time-to-event prediction problems by modeling the survival function as a weighted mixture of individual parametric survival distributions. \cite{nagpal2021deepcox} further extended their work to a mixture of Cox model, by assuming there exists a latent group structure and within each group, the proportional hazards assumption holds. \cite{zhong2021deep} considered deep neural networks to generalize the extended hazard model, which includes both the Cox proportional hazards model and the accelerated failure time model as special cases, and theoretically proved the consistency as well as the rate of convergence of the proposed estimator. However, all the works mentioned above failed to touch the variable selection problem in survival, and the main objective of this paper is to fill the gap.

\section{Methods} \label{Method}
\begin{figure}[ht!]
  \centering
  \includegraphics[width=3in]{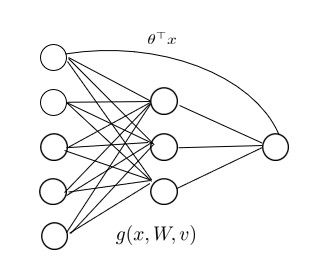}
  \caption{\it The architecture of LassoNet. $\bm{\theta}^{\top}\bm{x}$ is a residual layer and $g(\bm{x}, \bm{W}, \bm{v})$ is a feed-forward neural network.}
  \label{figure:architecture}
\end{figure}

Before introducing our method, let us briefly review the classical penalized partial likelihood techniques for Cox proportional hazards model. To the best of our acknowledge, \cite{tibshirani1996regression} is the first attempt to use the penalization method to achieve variable selection on survival data, where the author adds an $\ell_1$ penalty term in the partial likelihood function to estimate $\bm{\beta}$, that is,
\begin{equation}
\hat{\bm{\beta}}_{\textrm{lasso}} = \argmin -\ell(\bm{\beta}), \textrm{ subject to } \sum_{j=1}^p|\beta_j| \leq s,
\end{equation}
where $s$ is a tuning parameter. The above equation can be rewritten equivalently in a Lagrangian form
\begin{equation}
\hat{\bm{\beta}}_{\textrm{lasso}} = \argmin -\ell(\bm{\beta}) + \lambda\sum_{j=1}^pJ(\beta_j),
\end{equation}
where $J(\beta_j) = |\beta_j|$ and the exact relationship between $s$ and $\lambda$ is data dependent. In addition to the $\ell_1$ penalty, some other penalty forms have also been studied in the literature, such as adaptive Lasso (\cite{zhang2007adaptive}), SCAD (\cite{fan2002variable}). However, the aforementioned variable selection methods ignore feature interaction and nonlinear structure, to overcome the problem, we extend the advantages of LassoNet (\cite{LassoNet}) to survival data to tackle nonlinearity and complex structure. Before presenting the proposed variable selection approach using LassoNet, we first briefly review the notations and terminologies used in deep neural networks. Unlike traditional approximation theory in mathematics where people approximate complex functions based upon the summation of a series of simple functions, such as the Taylor Series and Fourier series, deep learning uses the compositions of simple functions. To be more specific, from a high-level point of view, a typical feed-forward neural network $g(\bm{x}, \bm{W}, \bm{v})$ is a composition of several simple functions, that is,
\begin{equation}\label{fnn}
g(\bm{x}, \bm{W}, \bm{v}):=\bm{W}_{L}\sigma_{\bm{v}_{L}} \ldots \bm{W}_1 \sigma_{\bm{v}_1} \bm{W}_0 \bm{x}, \quad \bm{x}\in\bbR^{d_0}, d_0 = p,
\end{equation}
where $\bm{W}\in\cW:=\{(\bm{W}_0,\ldots,\bm{W}_{L}): \bm{W}_{l}\in \mathbb{R}^{d_{l+1}\times d_{l}}, 0\le l\le L\}$ is the weight matrix, $\bm{v}\in\cV:=\{(\bm{v}_1,\ldots,\bm{v}_{L}): \bm{v}_{l}\in \mathbb{R}^{d_l}, 1\le l\le L\}$ is the bias term, and $\sigma_{\bm{v}}: \mathbb{R}^r \rightarrow \mathbb{R}^r$ is a non-linear activation function to learn the complex pattern from the data. Here $L$ is the number of hidden layers, i.e., the length of the network, and $\bm{d}=(d_0=p, \ldots, d_{L+1}), d_{L+1}=1, d_j>0$ is the number of units in each layer, i.e., the depth of the network. In our paper, we consider the following shift activation function
\begin{equation*}
\sigma_{\mathbf{v}}\left(\begin{array}{c}
y_1 \\
\vdots \\
y_r
\end{array}\right)=\left(\begin{array}{c}
\sigma\left(y_1-v_1\right) \\
\vdots \\
\sigma\left(y_r-v_r\right)
\end{array}\right),
\end{equation*}
where $\bm{v}=(v_1,\ldots, v_r)\in\mathbb{R}^r$. The parameters need to be estimated from the data are $(\bm{W}_j)_{j=0,\ldots,L}$ and $(\bm{v}_j)_{j=1,\ldots,L}$. In practice, the network depth is of crucial importance to capture the complex hidden structure of the data. A deeper network is able to learn a more complex representation of the input data, however, with the network depth increasing, accuracy gets saturated and then degrades rapidly, this is called degradation problems. To overcome this problem, \cite{Literature_Review_CV} introduced a famous framework, denoted as residual neural networks. By adding skip connections, the number of layers in residual neural networks can easily achieve more than one hundred. Thus, we consider the following class of residual feed-forward neural networks
\begin{equation}\label{res_net}
\mathcal{F} = \{f(\bm{x}, \bm{W}, \bm{v}, \bm{\theta}):=\bm{\theta}^\top\bm{x} + g(\bm{x}, \bm{W}, \bm{v}), \quad g(\bm{x}, \bm{W}, \bm{v}) \textrm{ of form (\ref{fnn})}\}, \quad \bm{\theta}, \bm{x}\in\bbR^{p}.
\end{equation}
To tackle the complicated structure between the covariates and survival time, we replace the linear function in (\ref{CoxPH}) by a complex nonlinear function $\Phi(\bm{x}, \bm{W}, \bm{v}, \bm{\theta}) \in \mathcal{F}$ and the corresponding hazard function and partial likelihood function now becomes
\begin{align}\label{CoxPH_nonparametric}
h(t|\bm{x}) &= h_0(t)\exp[\Phi(\bm{x}, \bm{W}, \bm{v}, \bm{\theta})],\\
 L(\bm{W}, \bm{v}, \bm{\theta}) &= \prod_{i=1}^{n}\left(\frac{\exp[\Phi(\bm{x}, \bm{W}, \bm{v}, \bm{\theta})]}{\sum_{j\in \mathcal{R}(Y_i)}\exp[\Phi(\bm{x}, \bm{W}, \bm{v}, \bm{\theta})]}\right)^{\delta_i}.
\end{align}
By replacing the linear relationship $\bm{x}^\top\bm{\beta}$ to $\Phi(\bm{x}, \bm{W}, \bm{v}, \bm{\theta})$, the approach is capable of modeling interactions and nonlinearity between the covariates and survival time. In order to achieve variable selection, penalty terms are needed in the following optimization problem

\begin{align}\label{fnn.variable.selection}
(\widehat{\bm{W}}, \widehat{\bm{v}}, \widehat{\bm{\theta}}) &= \argmin L(\bm{W}, \bm{v}, \bm{\theta})+ \lambda\|\bm{\theta}\|_1,\\
 &\textrm{ subject to } \|W_{0,i}\|_\infty \leq M|\theta_i|, i = 1, \ldots, p.
\end{align}
In \eqref{fnn.variable.selection}, $\theta_i$ is the $i$-component of $\bm{\theta}\in\bbR^{p}$ and $W_{0,i}$ contains the weights  for the $i$-th input variable in the first hidden layer. There are two tuning parameters in the objective function: $\lambda$ and $M$, which penalize the linear and nonlinear components simultaneously. The constraint $\|W_{0,i}\|_\infty \leq M|\theta_i|$ plays a vital role, in that $M$ leverages the effect of the $i$-th input variable, thereby capturing the non-linearity in the data. When $M=0$, the neural network part $g(\bm{x}, \bm{W}, \bm{v})$ in $\Phi(\bm{x}, \bm{W}, \bm{v}, \bm{\theta})$ vanishes and $\Phi(\bm{x}, \bm{W}, \bm{v}, \bm{\theta}) = \bm{\theta}^{\top}\bm{x}$, and thus \eqref{fnn.variable.selection} degenerates to standard LASSO in (\cite{tibshirani1996regression}). When $M \rightarrow \infty$, we get a feed-forward network with an $\ell_1$ penalty on the residual layer.  Following the optimization procedure outlined in \cite{LassoNet}, we optimize the objective function \eqref{fnn.variable.selection} using proximal gradient descent. It is worth mentioning that without domain knowledge, it is difficult to determine the value of $M$.  In practical implementation, we use cross-validation to select $M$, a common practice in machine learning algorithms. By introducing the residual layer $\bm{\theta}^\top\bm{x}$, the linear and non-linear components can be optimized simultaneously. The architecture of the network and the detailed pseudocode are summarized in Figure \ref{figure:architecture} and Algorithm \ref{algo2}.

\begin{algorithm}
\caption{}
\label{algo2}
\begin{algorithmic}
\State \textbf{Input}: observed data set is $\big\{(Y_i, \delta_i, \bm{x}_i): Y_i \geq 0, \delta_i \in \{0, 1\}, \bm{x}_i \in \mathbb{R}^p, i=1,\ldots,n \big\}$, residual network $\Phi(\bm{x}, \bm{W}, \bm{v}, \bm{\theta}) \in \mathcal{F}$, number of epochs $E$,  learning rate $\alpha$, hyper-parameter $M$, and path multiplier $\epsilon$
\State Initialize $\bm{\theta}, \bm{W}, \bm{v}$, $\lambda, k=p$, and loss function $L(\bm{W}, \bm{v}, \bm{\theta}) = \prod_{i=1}^{n}\left(\frac{\exp[\Phi(\bm{x}, \bm{W}, \bm{v}, \bm{\theta})]}{\sum_{j\in \mathcal{R}(Y_i)}\exp[\Phi(\bm{x}, \bm{W}, \bm{v}, \bm{\theta})]}\right)^{\delta_i}$
\While{$k > 0$}
\State Update $\lambda \leftarrow (1+\epsilon)\lambda$
\For{$e \in \{1, \ldots, E\}$}
\State Compute gradients $\nabla _{\bm{\theta}} \ell, \nabla _W \ell, \nabla _v \ell$ using back-propagation
\State Update $\bm{\theta} \leftarrow \bm{\theta} - \alpha\nabla _{\bm{\theta}}\ell, W \leftarrow W - \alpha\nabla _W \ell, v \leftarrow v - \alpha\nabla _v\ell$
\State Update $(\bm{\theta}, W_0) \leftarrow \textrm{Hier-Prox}(\theta, W_0, \alpha\lambda, M)$, where \State Hier-Prox is provided in the Algorithm 2 in \cite{LassoNet}
\EndFor
\State Update $k$ to be the number of non-zero elements of $\bm{\theta}$
\EndWhile
\end{algorithmic}
\end{algorithm}
\section{Numerical Results} \label{Simulation}
In this section, we carry out simulation studies to assess the finite sample performance of the LassoNet on right-censored survival data. In both designs, we use the same network architecture with length $L = 3$ and width $N = 30$. The dropout probability is set to 0.2 to avoid overfitting. Summary statistics from each simulation setting are calculated based on 100 independent simulation runs. In each setting, we use the inverse probability method by \cite{bender2005generating} to generate the event times from the hazard function. Suppose the conditional survival function under Cox proportional hazards model in (\ref{CoxPH}) is
\begin{equation*}
S(t|\bm{x}) = \exp\left(-H_0(t)\exp(\psi(\bm{x}))\right),
\end{equation*}
where $H_0(t)$ is the cumulative baseline hazard function and $\psi(\bm{x})$ is the covariates effect. let $U$ be uniformly distributed on $[0, 1]$, then the corresponding event time
\begin{equation}\label{generate_T_from_hazard}
T = S^{-1}(U|\bm{x}) = H_0(t)^{-1}\left(-\frac{\log(U)}{\exp(\psi(\bm{x}))}\right).
\end{equation}
In order to evaluate the performance of the variable selection method, we consider the following three metrics: 
\begin{itemize}
\item MinSize = the minimum number of selected variables to includes all true variables
\item $\textrm{Prob}(k, \textrm{all})$ = the success rate that the selected top $k$ variables are all true variables
\item $\textrm{Prob}(k, i)$ = the success rate that the selected top $k$ variables contain variable $x_i$
\end{itemize}
These three metrics are widely used in feature selection literature. Suppose the number of true variables is $k$, then by definition, MinSize is expected to be at least $k$; and the closer to $k$, the better the procedure. The metric $\textrm{Prob}(k, \textrm{all})$ measures the sensitivity (true positive) rate of the method in detecting the true variables. A higher value of $\textrm{Prob}(k, \textrm{all})$ is desirable, as this indicates there is a higher chance that the algorithm will pick all the true variables. The metric $\textrm{Prob}(k, i)$ measures the true positive and false positive rate of the proposed method. If the set of the true variable contains variable $x_i$, then the higher value of $\textrm{Prob}(k, i)$, the better of the model, in contrast, if $x_i$ is not associated with the survival time, then the lower value of $\textrm{Prob}(k, i)$, the better of the model.

We compare the proposed model $\textrm{Cox}_\textrm{LassoNet}$ with three other methods: (1) classical Cox proportional hazards model $\textrm{Cox}_\textrm{classical}$; (2) $\ell_1$ regularized Cox model $\textrm{Cox}_\textrm{Lasso}$; and (3) stepwise variable selection Cox proportional hazards model $\textrm{Cox}_\textrm{Stepwise}$. In each method, we rank the importance of each feature and select the top $k$ features accordingly. The importance of each feature from model $\textrm{Cox}_\textrm{classical}$ is based on the $p$-value associated with each variable. The $\ell_1$ regularized Cox model $\textrm{Cox}_\textrm{Lasso}$ and stepwise variable selection Cox proportional hazards model $\textrm{Cox}_\textrm{Stepwise}$ are implemented by the ``\textbf{coxphMIC}'' package and ``\textbf{pec}'' package in R.

\subsection{Model 1: Linear Cox proportional hazards model}
\begin{figure}[ht!]
  \centering
  \includegraphics[width=5in]{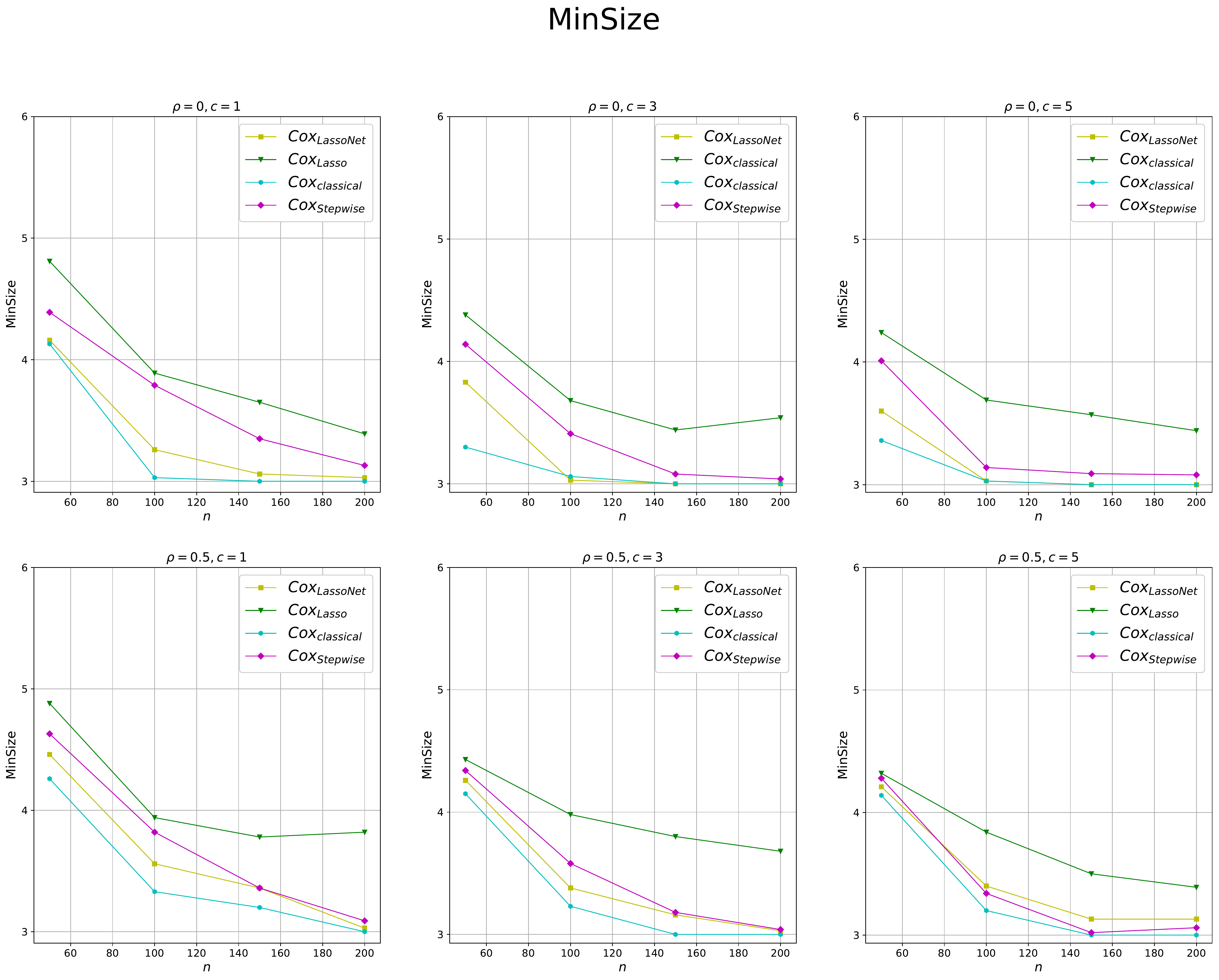}
  \includegraphics[width=5in]{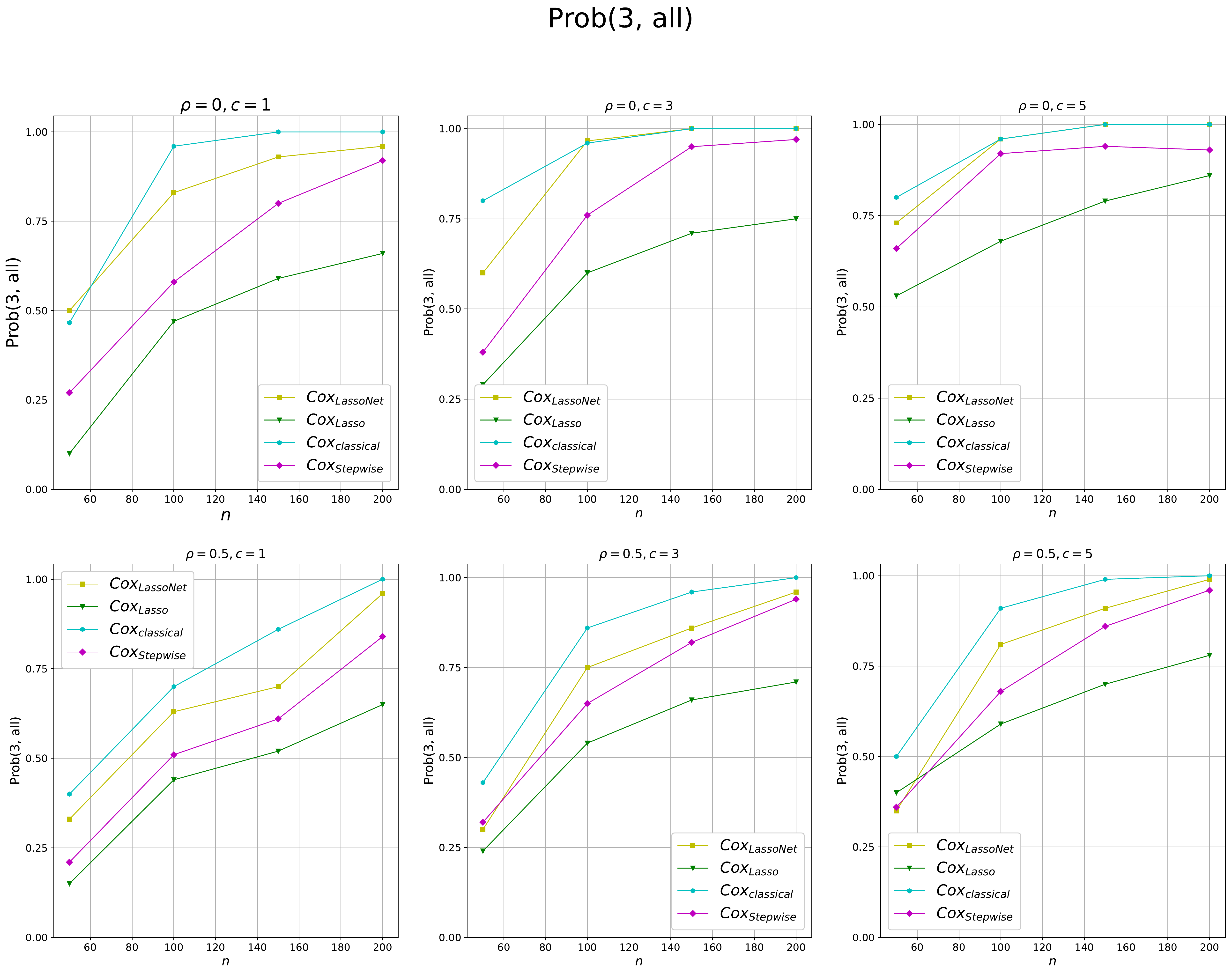}
  \caption{\it MinSize and Prob(3, all) of Simulation 1. The true hazard function is a linear function defined as $h(t) = 0.8x_1 + x_4 + 0.6 x_9$. LassoNet: $\textrm{Cox}_\textrm{LassoNet}$; Lasso: $\textrm{Cox}_\textrm{Lasso}$; Cox: $\textrm{Cox}_\textrm{classical}$; Piecewise: $\textrm{Cox}_\textrm{Stepwise}$}
  \label{figure:Sim_1_MinSize_ProbAll}
\end{figure}

\begin{figure}[ht!]
  \centering
  \includegraphics[width=5in]{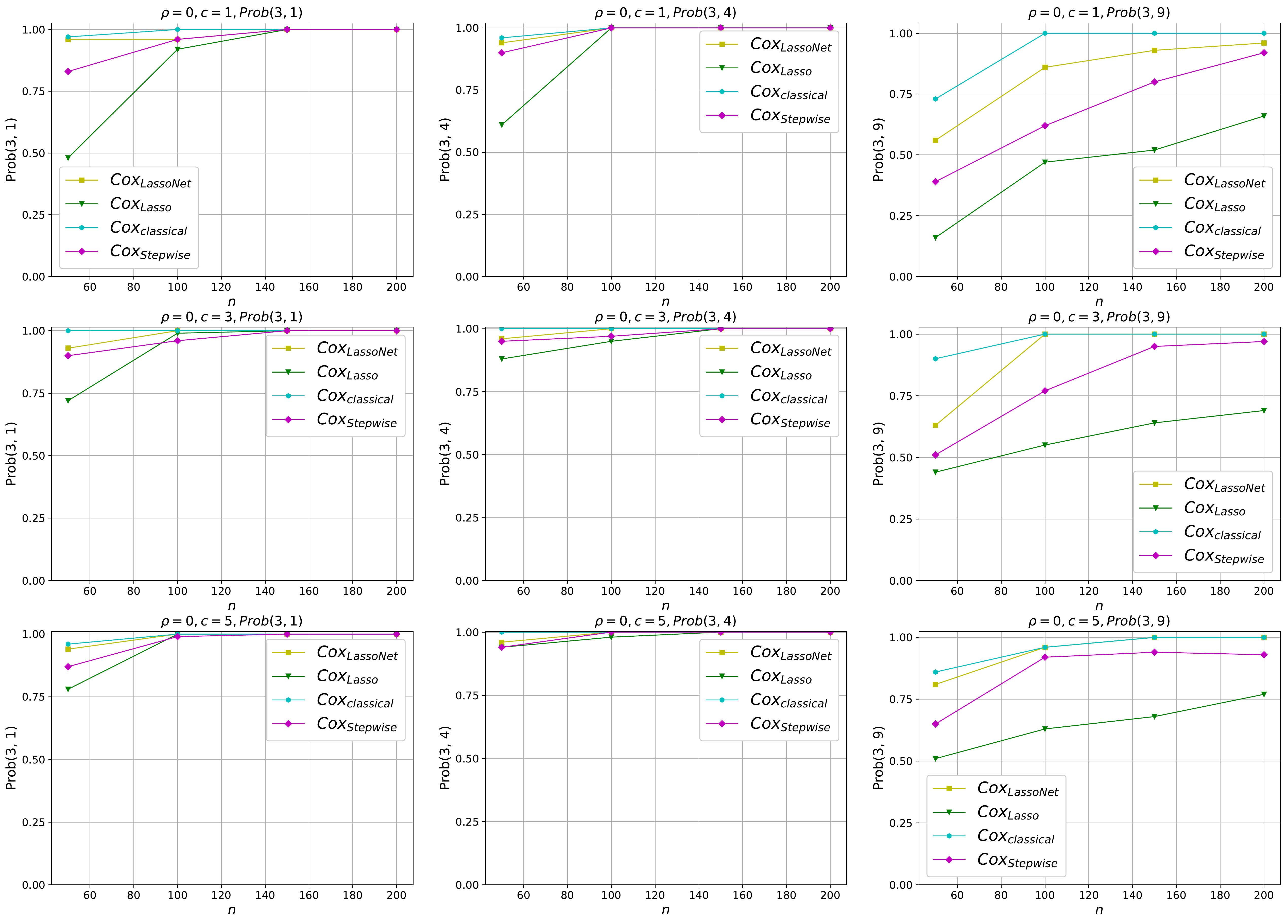}
  \caption{\it Prob(3, 1), Prob(3, 4), and Prob(3, 9) of Simulation 1. The true hazard function is a linear function defined as $h(t) = 0.8x_1 + x_4 + 0.6 x_9$.}
  \label{figure:Sim_1_ProbSingle}
\end{figure}

In the first model, we consider the classical Cox proportional hazards model in (\ref{CoxPH}) with $\bm{\beta} = (0.8, 0, 0, 1, 0, 0, 0, 0, 0.6, 0)^\top$, and the ten covariates $\bm{x} = (x_1, \ldots, x_{10})$ are from multivariate Gaussian distribution with piecewise correlation $\textrm{Cor}(x_i, x_j) = \rho^{|i-j|}, i,j=1,\ldots, 10$. Event times are generated from (\ref{generate_T_from_hazard}) with $\psi(\bm{x}) = \bm{x}^\top\bm{\beta}$ and without loss of generality, we let $h_0(t) = 1$. Censoring times $C$ are generated from a uniform distribution $U(0, c)$, where $c$ is a predetermined parameter to control the rates of censored samples. The MinSize and $\textrm{Prob}(3, \textrm{all})$ are summarized in Figure \ref{figure:Sim_1_MinSize_ProbAll}, which suggests that the MinSize decreases and $\textrm{Prob}(3, \textrm{all})$ increases as $n$ increases. Moreover, as $c$ increases, the results get better. This is because increasing $c$ will diminish the impact of censoring. Furthermore, we investigate the influence of $\rho$, i.e., the correlation of the covariates. We observe that when the dependence increases ($\rho$ increases), it is more difficult to select the true variables for all the methods. From Figure \ref{figure:Sim_1_MinSize_ProbAll}, we also find that classical Cox regression outperforms other methods. This is mainly because the data generating process is nothing but the linear Cox proportional hazards model thus the parametric test is more powerful than other nonparametric methods. Among all the other three methods $\textrm{Cox}_\textrm{LassoNet}$, $\textrm{Cox}_\textrm{Lasso}$, and $\textrm{Cox}_\textrm{Stepwise}$, the deep learning method is the best. The results of $\textrm{Prob}(k, i), i=1, 4, 9$ is reported in Figure \ref{figure:Sim_1_ProbSingle} to check the probability of selecting each true variable for each method. It is straightforward to see that, $\textrm{Prob(3, 4)} \geq \textrm{Prob(3, 1)} \geq \textrm{Prob(3, 9)}$, which is consistent with the fact that the coefficient of $x_4$ is greater than $x_1$ and $x_9$. We only report the results for $\rho=0$ to save space, and the pattern for $\rho=0.5$ is similar.
\subsection{Model 2: Nonlinear Cox proportional hazards model}
\begin{figure}[ht!]
  \centering
  \includegraphics[width=5in]{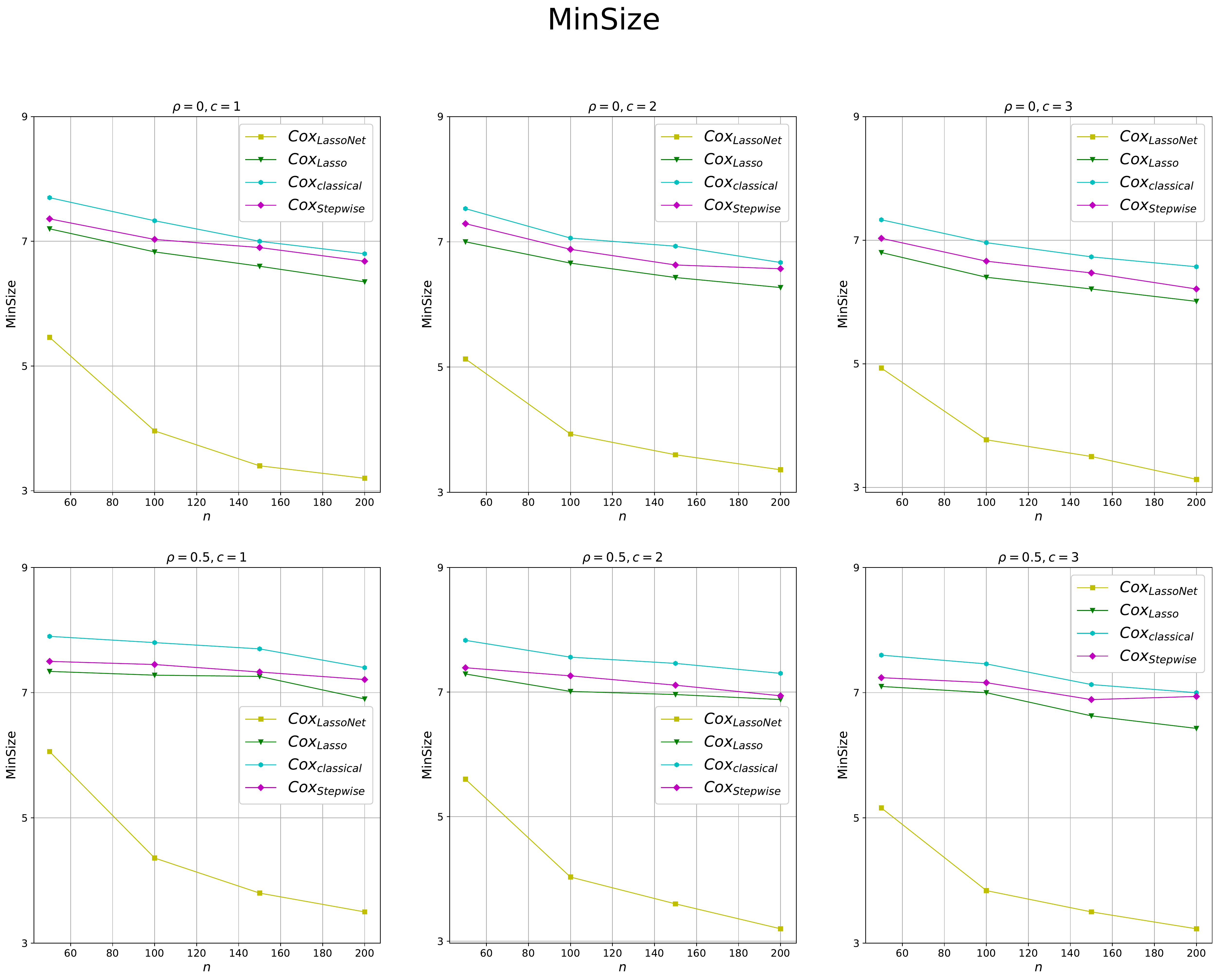}
  \includegraphics[width=5in]{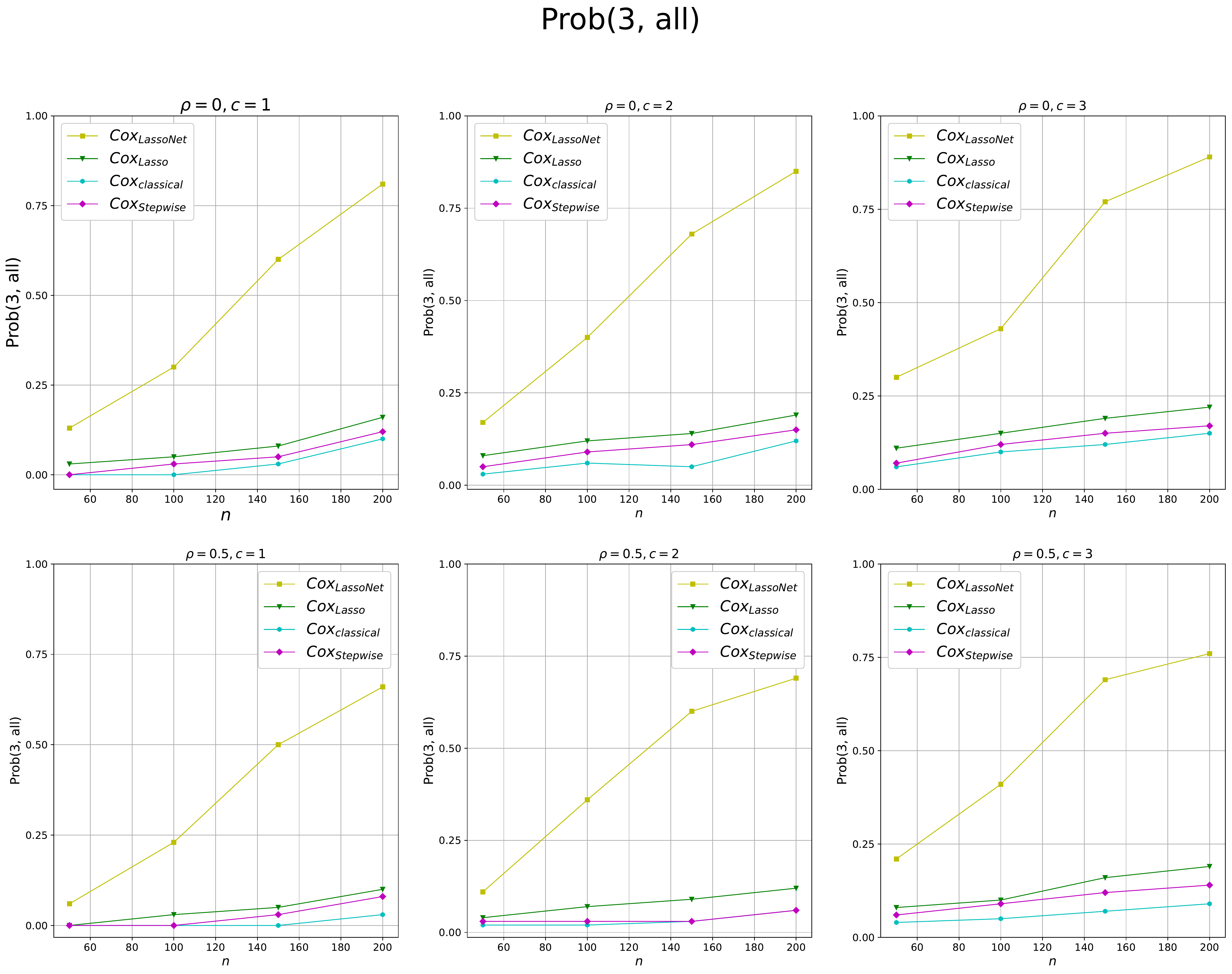}
  \caption{\it MinSize and Prob(3, all) of Simulation 2. The true hazard function is a nonlinear function defined as $h(t) = x_1^2 + \max(x_4, 1) + x_4\times x_9$.}
  \label{figure:Sim_2_MinSize_ProbAll}
\end{figure}

\begin{figure}[ht!]
  \centering
  \includegraphics[width=5in]{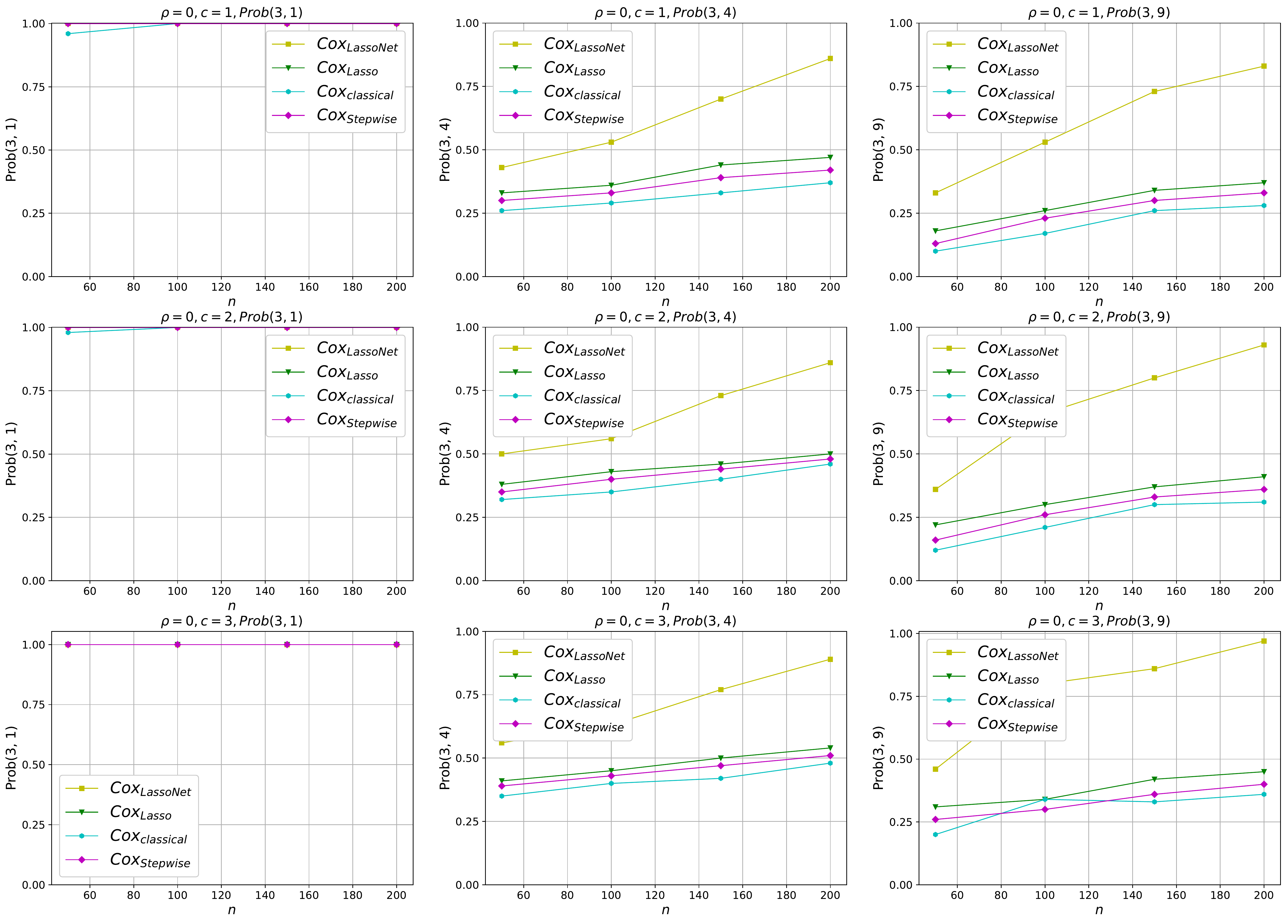}
  \caption{\it Prob(3, 1), Prob(3, 4), and Prob(3, 9) of Simulation 2. The true hazard function is a nonlinear function defined as $h(t) = x_1^2 + \max(x_4, 1) + x_4\times x_9$.}
  \label{figure:Sim_2_ProbSingle}
\end{figure}

In the second example, we consider a nonlinear Cox proportional hazards where the hazard function is generated by
\begin{equation*}
h(t) = x_1 + \max(x_4, 1) + x_4\times x_9.
\end{equation*}
Similar to Simulation 1, we generate the ten covariates $\bm{x} = (x_1, \ldots, x_{10})$ are from multivariate Gaussian distribution with piecewise correlation $\textrm{Cor}(x_i, x_j) = \rho^{|i-j|}, i,j=1,\ldots, 10$ and the baseline hazard function $h_0(t) = 1$. In this example, variable $x_1$ is a linear term while $x_4, x_9$ are nonlinear and interacted terms. The goal of this setting is to compare the ability of selecting the nonlinear variables. The summary statistics of different metrics are summarized in Figure \ref{figure:Sim_2_MinSize_ProbAll} and Figure \ref{figure:Sim_2_ProbSingle}. It is straightforward to see that LassoNet is more capable of selecting the nonlinear features than the other three methods. It is worth mentioning that $\textrm{Prob(3, 1)}$ is very closed to $1$ for all the methods, however, $\textrm{Prob(3, 4)}$ and $\textrm{Prob(3, 9)}$ are very low for all the other methods.



\section{Real Data Analysis} \label{Real_Data}
In this section, we apply the proposed methods to the diffuse large-B-cell lymphoma data (\cite{realdata}). The data set consists of n = 240 observations with diffuse large-B-cell lymphoma after chemotherapy and 138 patients die during the follow-up study, that is, the censor rate is $138/240=57.5\%$. DNA microarrays techniques were examined for gene expression to analyze for genomic abnormalities. The purpose of this analysis is to identify the prognostic genes among $p=7399$ microarray features. The length and width of the network are $L=3, N=30$, respectively with a dropout rate equal $0.2$.

We standardize each predictor to have amean zero and standard deviation $1$ and apply LassoNet to select the top five genes, which are: BC012161, LC33732, X00457, AK000978, and LC24433. It is interesting that three of the selected genes, BC012161, X00457, and LC24433 belong to the proliferating cells, major-histocompatibility-complex class, and immune cells in the lymph node, respectively. These three classes were the gene-expression signatures defined in \cite{realdata}. The other two selected genes do not belong to the classes defined in \cite{realdata}, however, they may also be related to lymphoma as reported by \cite{sha2006bayesian}. To further verify the selected genes are significant, we fit the Cox proportional hazard model with all these five genes and report the $p$-values. The $p$-values are all less than $10^{-4}$ which confirms that the five genes are significant. The results are summarized in Table \ref{realdata}.

\begin{table}[]
\caption{Top five genes selected by LassoNet for diffuse large-B-cell lymphoma data.}
\label{realdata}
\centering
\begin{tabular}{lll}
Genes    & Description & $p$-value \\ \hline
BC012161 &   reported by \cite{realdata}          &    $<10^{-5}$       \\
LC33732  &   reported by \cite{sha2006bayesian}   &    $<10^{-5}$       \\
X00457   &   reported by \cite{realdata}          &    $<10^{-5}$       \\
AK000978 &   reported by \cite{sha2006bayesian}   &    $<10^{-5}$       \\
LC24433  &   reported by \cite{realdata}          &    $<10^{-4}$       \\\hline
\end{tabular}
\end{table}

\section{Conclusion} \label{Conclusion}
The successful application of deep neural networks in variable selection problem on survival data is limited, In this work, we extend the state-of-the-art deep learning variable selection algorithm to survival data. Owing to the superiority of deep neural networks to capture complex structure feature space, the proposed method is able to select the informative features when the true hazard function is nonlinear. Simulation results verify the validity of the method and the real world data analysis indicates that the deep learning method can also be applied to survival data with the dimensionality that can be much larger than sample size.

\section*{Acknowledgement}
The authors are grateful for Peng Sun and Richard Foster for their many critical and constructive comments and suggestions that greatly improved the paper.
\clearpage
\bibliographystyle{apalike}
\bibliography{reference}

\begin{thebibliography}{}

\bibitem[Abid et~al., 2019]{abid2019concrete}
Abid, A., Balin, M.~F., and Zou, J. (2019).
\newblock Concrete autoencoders for differentiable feature selection and
  reconstruction.
\newblock {\em arXiv preprint arXiv:1901.09346}.

\bibitem[Bahdanau et~al., 2014]{Literature_Review_NLP}
Bahdanau, D., Cho, K., and Bengio, Y. (2014).
\newblock Neural machine translation by jointly learning to align and
  translate.
\newblock {\em arXiv preprint arXiv:1409.0473}.

\bibitem[Bender et~al., 2005]{bender2005generating}
Bender, R., Augustin, T., and Blettner, M. (2005).
\newblock Generating survival times to simulate cox proportional hazards
  models.
\newblock {\em Statistics in medicine}, 24(11):1713--1723.

\bibitem[Cox, 1972]{cox1972regression}
Cox, D.~R. (1972).
\newblock Regression models and life-tables.
\newblock {\em Journal of the Royal Statistical Society: Series B
  (Methodological)}, 34(2):187--202.

\bibitem[Du et~al., 2021]{du2021unified}
Du, M., Zhao, H., and Sun, J. (2021).
\newblock A unified approach to variable selection for cox’s proportional
  hazards model with interval-censored failure time data.
\newblock {\em Statistical Methods in Medical Research}, 30(8):1833--1849.

\bibitem[Fan et~al., 2010]{fan2010high}
Fan, J., Feng, Y., and Wu, Y. (2010).
\newblock High-dimensional variable selection for cox’s proportional hazards
  model.
\newblock In {\em Borrowing strength: Theory powering applications--a
  Festschrift for Lawrence D. Brown}, pages 70--86. Institute of Mathematical
  Statistics.

\bibitem[Fan et~al., 1997]{Fan1997}
Fan, J., Gijbels, I., and King, M. (1997).
\newblock {Local likelihood and local partial likelihood in hazard regression}.
\newblock {\em The Annals of Statistics}, 25(4):1661 -- 1690.

\bibitem[Fan and Li, 2001]{fan2001variable}
Fan, J. and Li, R. (2001).
\newblock Variable selection via nonconcave penalized likelihood and its oracle
  properties.
\newblock {\em Journal of the American statistical Association},
  96(456):1348--1360.

\bibitem[Fan and Li, 2002]{fan2002variable}
Fan, J. and Li, R. (2002).
\newblock Variable selection for cox's proportional hazards model and frailty
  model.
\newblock {\em The Annals of Statistics}, 30(1):74--99.

\bibitem[Faraggi and Simon, 1995]{faraggi1995neural}
Faraggi, D. and Simon, R. (1995).
\newblock A neural network model for survival data.
\newblock {\em Statistics in medicine}, 14(1):73--82.

\bibitem[Faraggi and Simon, 1998]{faraggi1998bayesian}
Faraggi, D. and Simon, R. (1998).
\newblock Bayesian variable selection method for censored survival data.
\newblock {\em Biometrics}, pages 1475--1485.

\bibitem[Han et~al., 2018]{han2018autoencoder}
Han, K., Wang, Y., Zhang, C., Li, C., and Xu, C. (2018).
\newblock Autoencoder inspired unsupervised feature selection.
\newblock In {\em 2018 IEEE international conference on acoustics, speech and
  signal processing (ICASSP)}, pages 2941--2945. IEEE.

\bibitem[He et~al., 2016]{Literature_Review_CV}
He, K., Zhang, X., Ren, S., and Sun, J. (2016).
\newblock Deep residual learning for image recognition.
\newblock In {\em Proceedings of the IEEE conference on computer vision and
  pattern recognition}, pages 770--778.

\bibitem[Ibrahim et~al., 1999]{ibrahim1999bayesian}
Ibrahim, J.~G., Chen, M.-H., and MacEachern, S.~N. (1999).
\newblock Bayesian variable selection for proportional hazards models.
\newblock {\em Canadian Journal of Statistics}, 27(4):701--717.

\bibitem[Ishwaran et~al., 2008]{Random_survival_forests}
Ishwaran, H., Kogalur, U.~B., Blackstone, E.~H., and Lauer, M.~S. (2008).
\newblock Random survival forests.
\newblock {\em The annals of applied statistics}, 2(3):841--860.

\bibitem[Jim{\'e}nez-Luna et~al., 2020]{Literature_Review_drug}
Jim{\'e}nez-Luna, J., Grisoni, F., and Schneider, G. (2020).
\newblock Drug discovery with explainable artificial intelligence.
\newblock {\em Nature Machine Intelligence}, 2(10):573--584.

\bibitem[Katzman et~al., 2018]{Deepsurv}
Katzman, J.~L., Shaham, U., Cloninger, A., Bates, J., Jiang, T., and Kluger, Y.
  (2018).
\newblock Deepsurv: personalized treatment recommender system using a cox
  proportional hazards deep neural network.
\newblock {\em BMC medical research methodology}, 18(1):1--12.

\bibitem[Kvamme et~al., 2019]{kvamme2019time}
Kvamme, H., Borgan, {\O}., and Scheel, I. (2019).
\newblock Time-to-event prediction with neural networks and cox regression.
\newblock {\em arXiv preprint arXiv:1907.00825}.

\bibitem[Lee et~al., 2018]{Deephit}
Lee, C., Zame, W., Yoon, J., and Van Der~Schaar, M. (2018).
\newblock Deephit: A deep learning approach to survival analysis with competing
  risks.
\newblock In {\em Proceedings of the AAAI conference on artificial
  intelligence}, volume~32.

\bibitem[Lemhadri et~al., 2021]{LassoNet}
Lemhadri, I., Ruan, F., Abraham, L., and Tibshirani, R. (2021).
\newblock Lassonet: A neural network with feature sparsity.
\newblock {\em Journal of Machine Learning Research}, 22(127):1--29.

\bibitem[Li et~al., 2021]{ODE}
Li, K., Wang, F., Liu, R., Yang, F., and Shang, Z. (2021).
\newblock Calibrating multi-dimensional complex ode from noisy data via deep
  neural networks.
\newblock {\em arXiv preprint arXiv:2106.03591}.

\bibitem[Li et~al., 2022]{DeepFS}
Li, K., Wang, F., and Yang, L. (2022).
\newblock Deep feature screening: Feature selection for ultra high-dimensional
  data via deep neural networks.
\newblock {\em arXiv preprint arXiv:2204.01682}.

\bibitem[Liu et~al., 2017]{liu2017deep}
Liu, B., Wei, Y., Zhang, Y., and Yang, Q. (2017).
\newblock Deep neural networks for high dimension, low sample size data.
\newblock In {\em IJCAI}, pages 2287--2293.

\bibitem[Luck et~al., 2017]{luck2017deep}
Luck, M., Sylvain, T., Cardinal, H., Lodi, A., and Bengio, Y. (2017).
\newblock Deep learning for patient-specific kidney graft survival analysis.
\newblock {\em arXiv preprint arXiv:1705.10245}.

\bibitem[Manduchi et~al., 2021]{manduchi2021deep}
Manduchi, L., Marcinkevi{\v{c}}s, R., Massi, M.~C., Weikert, T., Sauter, A.,
  Gotta, V., M{\"u}ller, T., Vasella, F., Neidert, M.~C., Pfister, M., et~al.
  (2021).
\newblock A deep variational approach to clustering survival data.
\newblock {\em arXiv preprint arXiv:2106.05763}.

\bibitem[Mirzaei et~al., 2020]{mirzaei2020deep}
Mirzaei, A., Pourahmadi, V., Soltani, M., and Sheikhzadeh, H. (2020).
\newblock Deep feature selection using a teacher-student network.
\newblock {\em Neurocomputing}, 383:396--408.

\bibitem[Nagpal et~al., 2021a]{nagpal2021deep}
Nagpal, C., Li, X., and Dubrawski, A. (2021a).
\newblock Deep survival machines: Fully parametric survival regression and
  representation learning for censored data with competing risks.
\newblock {\em IEEE Journal of Biomedical and Health Informatics},
  25(8):3163--3175.

\bibitem[Nagpal et~al., 2021b]{nagpal2021deepcox}
Nagpal, C., Yadlowsky, S., Rostamzadeh, N., and Heller, K. (2021b).
\newblock Deep cox mixtures for survival regression.
\newblock In {\em Machine Learning for Healthcare Conference}, pages 674--708.
  PMLR.

\bibitem[Ren et~al., 2019]{RNN}
Ren, K., Qin, J., Zheng, L., Yang, Z., Zhang, W., Qiu, L., and Yu, Y. (2019).
\newblock Deep recurrent survival analysis.
\newblock In {\em Proceedings of the AAAI Conference on Artificial
  Intelligence}, volume~33, pages 4798--4805.

\bibitem[Rosenwald et~al., 2002]{realdata}
Rosenwald, A., Wright, G., Chan, W.~C., Connors, J.~M., Campo, E., Fisher,
  R.~I., Gascoyne, R.~D., Muller-Hermelink, H.~K., Smeland, E.~B., Giltnane,
  J.~M., et~al. (2002).
\newblock The use of molecular profiling to predict survival after chemotherapy
  for diffuse large-b-cell lymphoma.
\newblock {\em New England Journal of Medicine}, 346(25):1937--1947.

\bibitem[Saeys et~al., 2007]{saeys2007review}
Saeys, Y., Inza, I., and Larranaga, P. (2007).
\newblock A review of feature selection techniques in bioinformatics.
\newblock {\em bioinformatics}, 23(19):2507--2517.

\bibitem[Sha et~al., 2006]{sha2006bayesian}
Sha, N., Tadesse, M.~G., and Vannucci, M. (2006).
\newblock Bayesian variable selection for the analysis of microarray data with
  censored outcomes.
\newblock {\em Bioinformatics}, 22(18):2262--2268.

\bibitem[Tibshirani, 1996]{tibshirani1996regression}
Tibshirani, R. (1996).
\newblock Regression shrinkage and selection via the lasso.
\newblock {\em Journal of the Royal Statistical Society: Series B
  (Methodological)}, 58(1):267--288.

\bibitem[Tibshirani, 1997]{tibshirani1997lasso}
Tibshirani, R. (1997).
\newblock The lasso method for variable selection in the cox model.
\newblock {\em Statistics in medicine}, 16(4):385--395.

\bibitem[Yi et~al., 2020]{yi2020simultaneous}
Yi, F., Tang, N., and Sun, J. (2020).
\newblock Simultaneous variable selection and estimation for joint models of
  longitudinal and failure time data with interval censoring.
\newblock {\em Biometrics}.

\bibitem[Zhang and Lu, 2007]{zhang2007adaptive}
Zhang, H.~H. and Lu, W. (2007).
\newblock Adaptive lasso for cox's proportional hazards model.
\newblock {\em Biometrika}, 94(3):691--703.

\bibitem[Zhao et~al., 2019]{zhao2019simultaneous}
Zhao, H., Wu, Q., Li, G., and Sun, J. (2019).
\newblock Simultaneous estimation and variable selection for interval-censored
  data with broken adaptive ridge regression.
\newblock {\em Journal of the American Statistical Association}.

\bibitem[Zhong et~al., 2021]{zhong2021deep}
Zhong, Q., Mueller, J.~W., and Wang, J.-L. (2021).
\newblock Deep extended hazard models for survival analysis.
\newblock {\em Advances in Neural Information Processing Systems}, 34.

\bibitem[Zhu et~al., 2017]{Zhu2017}
Zhu, X., Yao, J., Zhu, F., and Huang, J. (2017).
\newblock Wsisa: Making survival prediction from whole slide histopathological
  images.
\newblock In {\em Proceedings of the IEEE Conference on Computer Vision and
  Pattern Recognition}, pages 7234--7242.

\bibitem[Zou, 2006]{zou2006adaptive}
Zou, H. (2006).
\newblock The adaptive lasso and its oracle properties.
\newblock {\em Journal of the American statistical association},
  101(476):1418--1429.

\end{thebibliography}

\end{document}